\documentclass[10pt,twocolumn,letterpaper]{article}

\usepackage[pagenumbers]{iccv} % To force page numbers, e.g. for an arXiv version

\definecolor{iccvblue}{rgb}{0.21,0.49,0.74}
\usepackage[pagebackref,breaklinks,colorlinks,allcolors=iccvblue]{hyperref}

\def\paperID{5} % *** Enter the Paper ID here
\def\confName{ICCV}
\def\confYear{2025}

\title{MatSSL: Robust Self-Supervised Representation Learning for Metallographic Image Segmentation}

\author{
Hoang Hai Nam Nguyen$^{1,2\dagger}$,
Phan Nguyen Duc Hieu$^{1,2\dagger}$,
Ho Won Lee$^{1,2}$ \\
$^1$Korea Institute of Materials Science \\
$^2$University of Science and Technology \\
{\tt\small namhhn@ust.ac.kr, hieupnd@ust.ac.kr, h.lee@kims.re.kr} \\
{\small{$^{\dagger}$} Contribute equally}
}

\begin{document}
\maketitle
\begin{abstract}
MatSSL is a streamlined self-supervised learning (SSL) architecture that employs Gated Feature Fusion at each stage of the backbone to integrate multi-level representations effectively. Current micrograph analysis of metallic materials relies on supervised methods, which require retraining for each new dataset and often perform inconsistently with only a few labeled samples. While SSL offers a promising alternative by leveraging unlabeled data, most existing methods still depend on large-scale datasets to be effective. MatSSL is designed to overcome this limitation. We first perform self-supervised pretraining on a small-scale, unlabeled dataset and then fine-tune the model on multiple benchmark datasets. The resulting segmentation models achieve 69.13\% mIoU on MetalDAM, outperforming the 66.73\% achieved by an ImageNet-pretrained encoder, and delivers consistently up to nearly 40\% improvement in average mIoU on the Environmental Barrier Coating benchmark dataset (EBC) compared to models pretrained with MicroNet. This suggests that MatSSL enables effective adaptation to the metallographic domain using only a small amount of unlabeled data, while preserving the rich and transferable features learned from large-scale pretraining on natural images. The source code is fully available \footnote{The code is available at \url{https://github.com/duchieuphan2k1/MatSSL}}
\end{abstract}    
\section{Introduction}
\label{sec:intro}

Accurate microstructure segmentation underpins critical tasks in materials engineering from alloy design to failure analysis, yet remains challenging due to the extreme scarcity of pixel‐level annotations. Traditional supervised deep learning approaches \cite{nasa2022micronet, tutorialonthesegmentation}, even when leveraging transfer learning from ImageNet \cite{imagenet}, still suffer from inconsistent and low accuracy when trained on small, expert-labeled metallographic datasets, which are often limited to only a few dozen micrographs per material system.

Unsupervised and semi‐supervised pipelines (e.g., clustering, pseudo-labeling) can exploit unlabeled data \cite{balestriero2023cookbook,kim2020unsupervised}, but they typically cannot achieve the fine‐grained accuracy needed for reliable characterization. Self-supervised learning (SSL) offers an attractive alternative by using pretext tasks like contrastive, generative, predictive, or self-labeling to learn useful representations without ground‐truth masks \cite{chen2020simclr,he2019moco,rani2023sslreview,wen2025active}. Most state-of-the-art SSL models (e.g., MoCo, SimCLR) require pretraining on hundreds of thousands to millions of images from a related domain to achieve good performance. This poses a major challenge in metallography, where only a few thousand unlabeled micrographs are typically available.

In this work, we propose MatSSL, a simple and lightweight contrastive SSL framework tailored to metallographic images across diverse materials. MatSSL adapts a single ResNet-50 \cite{resnet} encoder using a novel Gated Feature Fusion mechanism without the dual encoders or large memory queues used in MoCo \cite{chen2020moco_v2, he2019moco} and SimCLR \cite{chen2020simclr}, and then transfers the adapted encoder into a U‐Net++ \cite{unet++} segmentation network for fine‐tuning on limited labeled data.

We evaluate MatSSL on multiple metallographic benchmark datasets: MetalDAM \cite{metaldam}, Aachen-Heerlen annotated steel microstructure dataset \cite{aachen}, and the Environmental Barrier Coating benchmark (EBC) few‐shot dataset \cite{nasa2022micronet}. The result of MatSSL demonstrates:
\begin{itemize}
  \item Achieving up to 3.2\% improvement in mIoU on MetalDAM compared to ImageNet‐pretrained baselines.
  \item Consistent gain of nearly 40\% average mIoU on EBC few‐shot dataset.
  \item Superior performance over standard SSL methods on small-scale metallographic datasets, surpassing MoCoV2 by up to 2.76\% on metallographic benchmarks.
\end{itemize}

This work shows that, with appropriate architectural modifications, SSL can be effective and practical even in extremely low-data regimes, enabling robust microstructure segmentation across a variety of metallurgical systems.

\section{Related Work}
\label{sec:related_work}
The segmentation of microstructures in metallic materials has been approached using a wide range of techniques. Early studies explored classical image processing methods such as thresholding and region growing \cite{gonzalez2006digital}, along with unsupervised approaches like clustering and traditional neural networks \cite{metaldam}. Despite their foundational contributions, these methods generally lack the robustness and accuracy achieved by modern deep learning models.
Convolutional Neural Networks (CNNs), trained on large annotated datasets, have shown significant improvements in identifying complex microstructural patterns. Enhancing segmentation performance in this context has required not only expanding the quantity and quality of labeled data \cite{fernandez2023tradeoff} or developing more sophisticated network architectures \cite{csunet}, but also integrating advanced training strategies. For instance, \cite{biswas2023microstructural} demonstrated that combining multiple color space transformations (RGB, HSV, YUV) can meaningfully improve segmentation accuracy, while transfer learning, data augmentation, and loss function optimizations (such as focal loss or polynomial decay) have also been leveraged to boost results. More recently, MicroNet \cite{nasa2022micronet}, supervised pretrained on a 100,000+ image microscopy corpus across 54 microstructural classes, shows that classifier pretraining on domain-specific data substantially boosts downstream few-class segmentation.

However, a persistent challenge in the field is the scarcity and inconsistent sharing of high-quality, pixel-level annotated datasets. Accurate annotation is both technically demanding and subjective, often requiring expert interpretation \cite{aachen}. Although unsupervised learning pipelines have been proposed to generate pseudo-labels \cite{kim2020unsupervised}, these typically still require human refinement to be viable for accurate segmentation.
Several public metallographic databases exist \cite{rozman2022dataset}, yet most focus on classification rather than segmentation or provide only partially annotated data \cite{aachen}. The Ultra High Carbon Steel (UHCS) database \cite{UHCSDB} is one of the largest available, offering over 900 samples, but only 24 have detailed segmentation annotations. MetalDAM \cite{metaldam} provides 42 annotated and 126 unannotated metallographic images. Despite these efforts, the abundance of unlabeled data compared to labeled data highlights the urgent need for methods that can leverage both to achieve robust segmentation. 

Self-supervised learning (SSL) \cite{densecl, chen2020moco_v2} has emerged as a powerful approach in domains facing similar data scarcity challenges, most notably in medical imaging. In this domain, the vast amount of available unlabeled data and the high cost of expert annotation have made SSL especially valuable for segmentation and diagnostic tasks. Comprehensive benchmarks \cite{kang2022benchmarking} have demonstrated that SSL pretraining on large collections of domain-specific, unlabeled images can yield models that outperform both ImageNet-pretrained and randomly initialized networks for various downstream tasks \cite{zbontar2021barlow, chen2020moco_v2, weinstein2013tcga, kim2024tulip}.

It is important to note, however, that nearly all SSL studies in medical imaging leverage large-scale unlabeled datasets for pretraining, often with tens of thousands or even hundreds of thousands of images. For example, \cite{kang2022benchmarking} uses datasets such as TCGA \cite{weinstein2013tcga}, CheXpert \cite{chexpert}, and MIMIC-CXR \cite{johnson2019mimiccxrjpglargepubliclyavailable}, each containing more than 100,000 images, while \cite{kalapos2022self} was pre-trained on thousands of MRI images. SSL frameworks like SimCLR and SimSiam (\cite{zbontar2021barlow}; \cite{chen2020moco_v2}) also depend on these vast resources to maximize representation learning.

In stark contrast, metallographic image analysis in materials science typically lacks access to such extensive datasets. The total number of available labeled images and unlabeled combined images is limited to just a few thousand. This fundamental difference poses unique challenges for applying SSL in materials microstructure segmentation and underscores the importance of developing SSL methods that are effective even in extremely low-data regimes. 
\section{Methodology}

\begin{figure}[t]
    \centering
    \includegraphics[width=\linewidth]{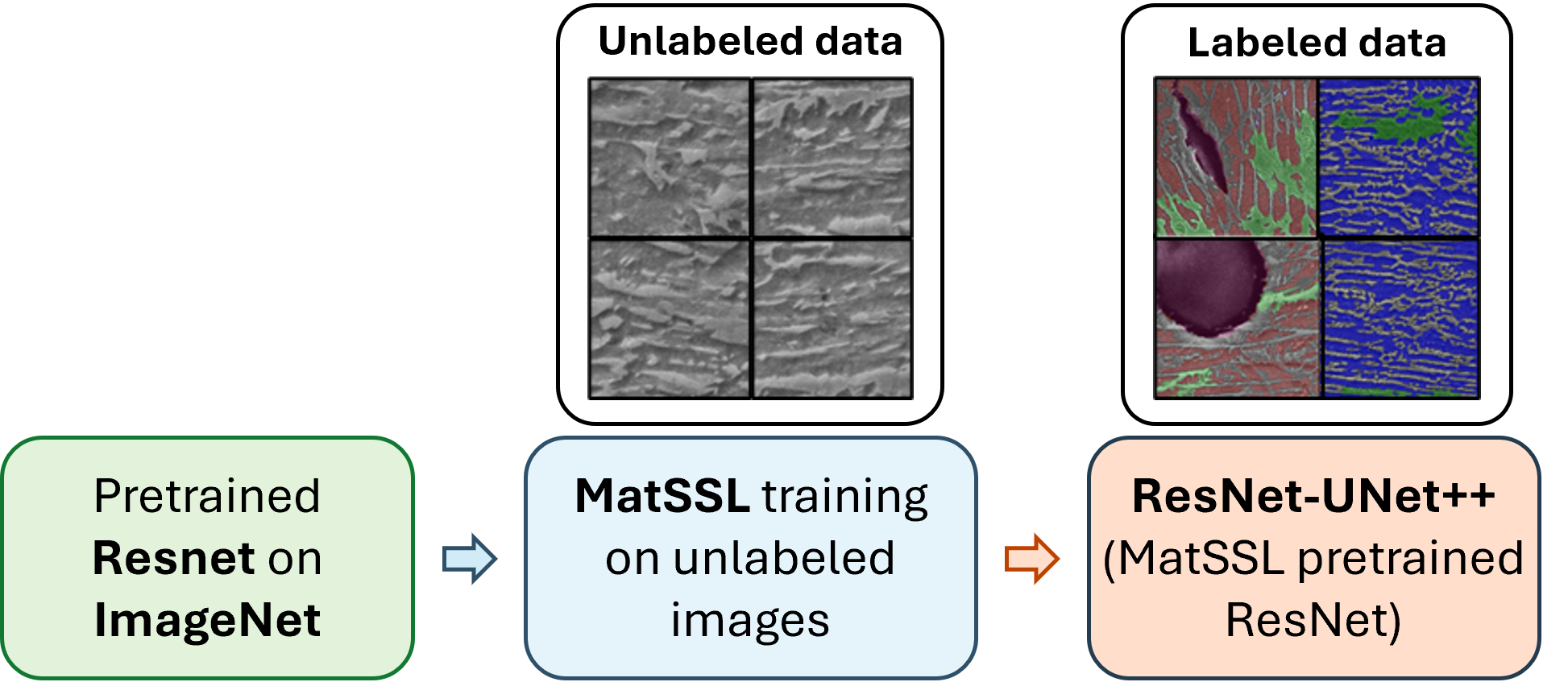}
    \caption{Overview of the training pipeline. MatSSL pretraining on unlabeled images is followed by supervised segmentation using U-Net++ with the pretrained ResNet encoder.}
    \label{fig:methodoverview}
\end{figure}

Our methodology involves two main steps, as shown in Fig.~\ref{fig:methodoverview}. First, we use a ResNet-50 backbone pretrained on ImageNet and further adapt it to the target domain via self-supervised learning (SSL) on unlabeled steel micrographs. We refer to this adapted encoder as MatSSL. Second, we plug the MatSSL pretrained encoder into a U-Net++ \cite{unet++} segmentation architecture and fine-tune end-to-end on labeled data.

\subsection{MatSSL Architecture}

\begin{figure*}[t]
    \centering
    \includegraphics[width=1.0\linewidth]{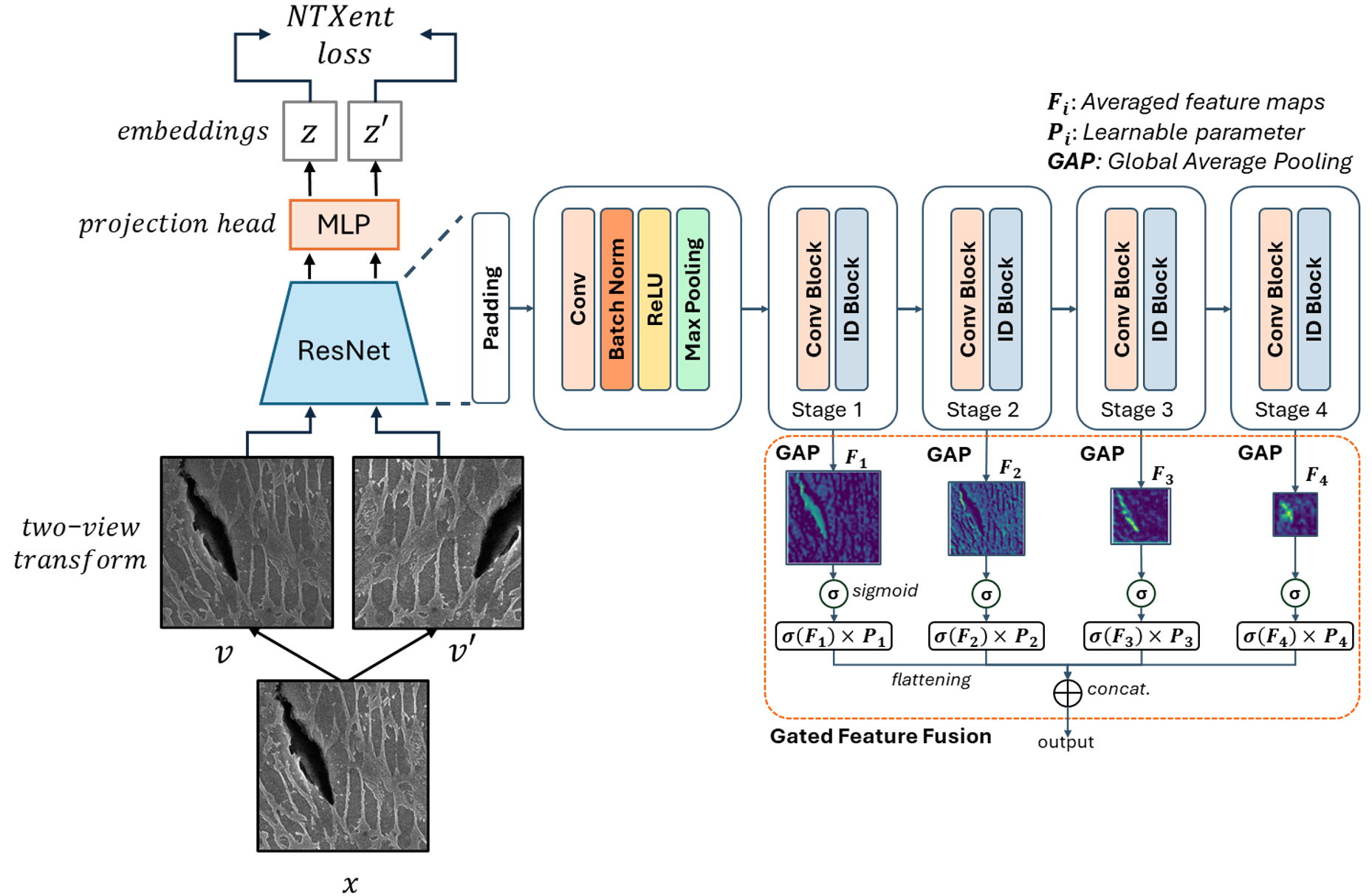}
    \caption{Overview of our proposed MatSSL architecture. Two augmented views of each input metallographic image are processed through a ResNet50 backbone. Gated Feature Fusion is applied at each stage to retain and adapt important pretrained features. The fused representations are passed through a projection head and optimized using the NTXentLoss in a contrastive learning framework.}
    \label{fig:MatSSL}
\end{figure*}

We employ a ResNet50 backbone in our self-supervised learning (SSL) framework to leverage pretrained ImageNet weights during SSL training. The SSL method used is contrastive learning, specifically utilizing the NTXentLoss \cite{ntxentloss}. Although this approach has proven effective in various tasks, applying standard contrastive learning directly to multi-stage backbones such as ResNet poses a significant drawback: essential features learned from millions of images during ImageNet pretraining can be lost as information flows through multiple sequential stages. This feature degradation becomes especially problematic in our setting, where the available SSL data for metallographic images is limited. As a result, if these pretrained features are not preserved, there can be a substantial drop in downstream segmentation accuracy, diminishing the effectiveness of SSL pretraining.

To address this issue, we modify the traditional contrastive learning architecture by introducing Gated Feature Fusion across the stages of ResNet. This mechanism connects intermediate features from each ResNet stage using learnable gates, aiming to retain and selectively update the crucial representations acquired during pretraining.  Specifically let \(F_i\) be the averaged feature map after Global Average Pooling (GAP) of all feature maps from the \(i\)-th ResNet stage, and \(P_i\) a learnable gating parameter. We compute a gated feature
\begin{equation}
\ G_i \;=\; \sigma(F_i)\;\times\;P_i
\label{eq:gate}
\end{equation}
where \(\sigma(\cdot)\) is the sigmoid activation. The \(\ G_i\) from all stages are concatenated and passed through a projection MLP to produce embeddings \(z\).

This gating mechanism allows the network to learn which features are important to retain, and which should be adapted to the metallographic domain, thereby minimizing the loss of valuable pretrained information as data passes through the backbone. After gated fusion, the updated features are passed through a projection head (as in SimCLR) before being used to compute the NTXentLoss for contrastive training of embedding \(z_i\) against its positive \(z_j\):

\begin{equation}
\ell_{i,j}
= -\log
\dfrac{
  \exp\bigl(\mathrm{sim}(z_i, z_j)/\tau\bigr)
}{
  \displaystyle
  \sum_{k=1}^{2N} \mathbf{1}_{[k\neq i]}\,
  \exp\bigl(\mathrm{sim}(z_i, z_k)/\tau\bigr)
},
\label{eq:ntxent}
\end{equation}

where \(\mathrm{sim}(\cdot,\cdot)\) is cosine similarity, \(\tau\) is the temperature, and \(2N\) is the batch size of contrastive pairs.

\subsection{Fine-tuning with Labeled Data}

After self-supervised pretraining, we integrate the MatSSL encoder into a U-Net++ \cite{unet++} semantic segmentation model. The decoder and segmentation head are randomly initialized, while the encoder weights come from the SSL stage. We then fine-tune the entire network on labeled metallographic images using the Dice loss:
\begin{equation}
\mathcal{L}_{\mathrm{dice}}
\;=\;
1 \;-\;
\frac{2 \sum_{i=1}^N y_i \,\hat y_i}
{\sum_{i=1}^N y_i \;+\; \sum_{i=1}^N \hat y_i},
\label{eq:dice}
\end{equation}
where \(y_i\) is the ground-truth (binary) label for pixel \(i\), \(\hat y_i\) its predicted probability, and \(N\) the total number of pixels.

This fine-tuning step leverages both domain-adapted features from SSL and the pixel-level annotations to produce accurate microstructure segmentations.

\section{Experiments}

\subsection{Datasets}
We explored these real‐world steel microstructure datasets:  
\begin{itemize}[leftmargin=*]
  \item UltraHigh Carbon Steel micrograph dataset \cite{UHCSDB} (refer as \textbf{UHCS}):\ 956 unlabeled SEM images of ultra‐high‐carbon steel microstructures.
  \item Aachen-Heerlen annotated steel microstructure dataset  \cite{aachen} (refer as \textbf{Aachen}):\ 1705 annotated images (classes: background, martensite/austenite).
  \item Metallography Dataset from Additive Manufacturing \cite{metaldam} (refer as \textbf{MetalDAM}):\ 178 unlabeled + 42 labeled images (4 classes: matrix, austenite, martensite/austenite, defect; 33 train / 9 test).
  \item Environmental-barrier Coatings benchmark dataset \cite{nasa2022micronet} (refer as \textbf{EBC}):\ We use 41 labeled images from EBC (environmental-barrier coatings) benchmark dataset (splits EBC-1–3).
\end{itemize}

\paragraph{Pre-training setup.}
We form SSL pretraining sets from unlabeled images in UHCS, MetalDAM, and labeled images in Aachen, excluding any images reserved for the test set of fine-tuning.  Each image was cropped into $256\times256$ patches with overlap rates of 0.6 (UHCS), 0.7 (MetalDAM), and 0.0 (Aachen), yielding 15,296, 14,566, and 16,836 patches, respectively.  These patches were combined into three SSL pretraining sets (Table \ref{tab:ssl-datasets}).

\begin{table}[h]
  \centering
  \caption{SSL pretraining sets (number of $256\times256$ patches).}
  \label{tab:ssl-datasets}
  \begin{tabular}{lc}
    \toprule
    Pretraining set & \# Patches \\
    \midrule
    UHCS + MetalDAM                  & 29.862 \\
    Aachen–Heerlen + UHCS            & 32.132 \\
    Aachen–Heerlen + UHCS + MetalDAM & 46.698 \\
    \bottomrule
  \end{tabular}
\end{table}

\paragraph{Fine-tuning setup.}
Downstream semantic segmentation is evaluated on the labeled splits of Aachen-Heerlen and MetalDAM, and on the EBC. For Aachen-Heerlen and MetalDAM, images are split into a train and test set with a ratio of 8:2, then are cropped into $256\times256$ patches without overlapping to preserve microstructural detail (see Table \ref{tab:finetune-datasets}). 

\begin{table}[h]
  \centering
  \caption{Train/test splits for Aachen and MetalDAM.}
  \label{tab:finetune-datasets}
  \begin{tabular}{lcc}
    \toprule
    Dataset & \# Train & \# Test \\
    \midrule
    Aachen & 1.403 & 302 \\
    MetalDAM       & 33    & 9   \\
    \bottomrule
  \end{tabular}
\end{table}

With EBC, we train and evaluate the model with the original splits and image size ($512\times512$) of the benchmark dataset, consisting of a train, val, and test set (see Table \ref{tab:finetune-datasets2}). We use val set to select the best model and compute accuracy on the totally unseen test set

\begin{table}[h]
  \centering
  \caption{Dataset splits on EBC.}
  \label{tab:finetune-datasets2}
  \begin{tabular}{lccc}
    \toprule
    Subset   & \# Train & \# Val & \# Test \\
    \midrule
    EBC-1   & 18 & 3 & 3 \\[-0.2em]
    EBC-2   & 4  & 3 & 3 \\[-0.2em]
    EBC-3   & 1  & 3 & 3 \\
    \bottomrule
  \end{tabular}
\end{table}

\subsection{Experimental Settings}
\label{sec:exp-settings}

\paragraph{Implementation.} 
All experiments were implemented in PyTorch v2.7.0+cu126 on an Intel system with two NVIDIA A100 GPUs. We fixed the random seed to 0 and enabled deterministic behavior for full reproducibility.

\paragraph{Data Preprocessing and Augmentations.} 
During \textbf{SSL Pretraining}, we follow the SimCLR \cite{chen2020simclr} augmentation pipeline: random resized crop, random horizontal flip, color jitter (brightness/contrast/saturation ±0.1), random grayscale conversion, Gaussian blur, and ImageNet normalization.

\paragraph{SSL Pre-training.} We train for 50 epochs with batch size 128 using SGD (initial lr=0.1, momentum=0.9, weight decay $10^{-6}$) and a cosine learning rate schedule decaying to $10^{-4}$. The NT-Xent loss uses a temperature \(\tau\)=0.07.

\paragraph{Fine-tuning and Evaluation.} The segmentation model U-Net++ implemented with \cite{iakubovskii2019segmentation} are fine-tuned using Adam (lr=$10^{-4}$, weight decay $10^{-5}$) for 200 epochs on MetalDAM and EBC, while trained with 50 epochs on Aachen. We fixed the batch size at 128 and recorded the mean Intersection-over-Union (mIoU).

\subsection{Results}

\begin{table*}[!t]
    \centering
    \caption{Comparison of mIoU (\%) for different pretraining strategies and finetune datasets.}
    \label{tab:main-results}
    \begin{tabular}{l l l c}
        \toprule
        Finetune Dataset & Pretrain & Pretrain Dataset & mIoU (\%) \\
        \midrule
        MetalDAM & super. ImageNet & - & 66.73 \\
                 & DenseCL & Aachen + UHCS & 68.76 \\
                 & MocoV2 & Aachen + UHCS & 67.18 \\
                 & \textbf{MatSSL} & \textbf{Aachen + UHCS} & \textbf{69.95} \\
                 & DenseCL & Aachen + UHCS + MetalDAM & 68.34 \\
                 & MocoV2 & Aachen + UHCS + MetalDAM & 68.40 \\
                 & \textbf{MatSSL} & \textbf{Aachen + UHCS + MetalDAM} & \textbf{69.02} \\
        \midrule
        Aachen   & super. ImageNet & - & 65.59 \\
                 & DenseCL & UHCS + MetalDAM & 65.82 \\
                 & MocoV2 & UHCS + MetalDAM & 65.90 \\
                 & \textbf{MatSSL} & \textbf{UHCS + MetalDAM} & \textbf{65.98} \\
                 & DenseCL & Aachen + UHCS + MetalDAM & 65.56 \\
                 & MocoV2 & Aachen + UHCS + MetalDAM & 65.65 \\
                 & \textbf{MatSSL} & \textbf{Aachen + UHCS + MetalDAM} & \textbf{65.86} \\
        \bottomrule
    \end{tabular}
\end{table*}

Table~\ref{tab:main-results} summarizes mean Intersection-over-Union (mIoU) for U-Net++ segmentation heads initialized with different pretrained encoders. Across both MetalDAM and Aachen–Heerlen, MatSSL consistently outperformed the ImageNet baseline and two popular SSL methods (DenseCL \cite{densecl}, MoCoV2 \cite{chen2020moco_v2}), especially with a small unlabeled dataset.

\begin{figure}[h]
  \centering
  \includegraphics[width=1.0\linewidth]{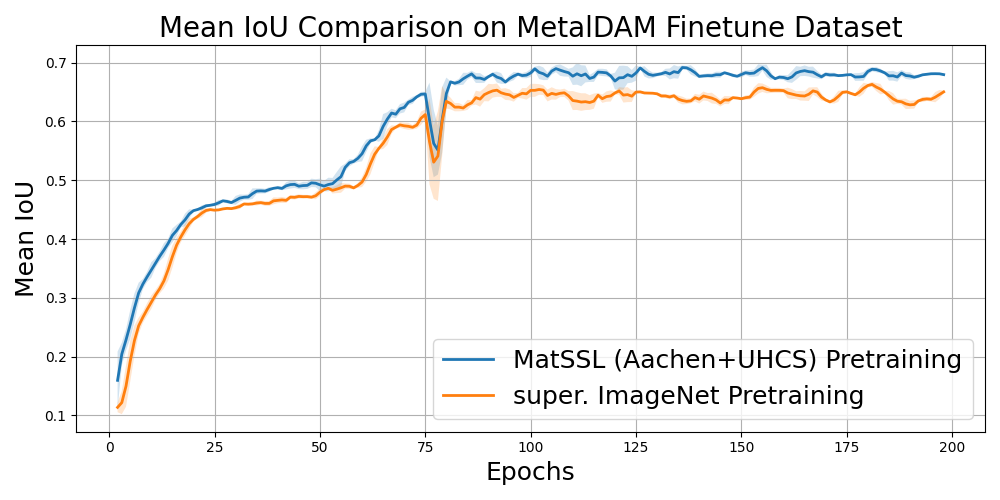}
  \caption{Validation mIoU per epoch on MetalDAM (Aachen+UHCS SSL vs.\ ImageNet).}
  \label{fig:uhcs-aachen-metaldam}
\end{figure}

\begin{figure}[h]
  \centering
  \includegraphics[width=1.0\linewidth]{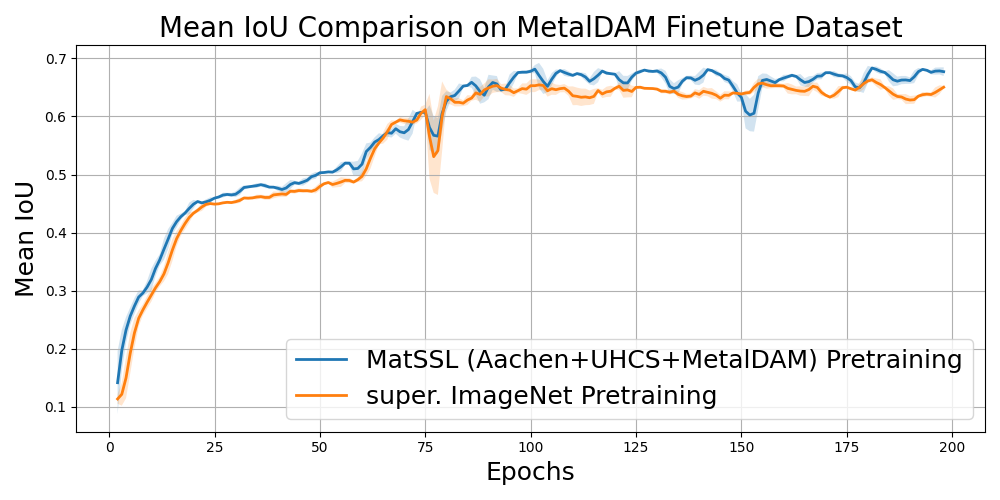}
  \caption{Validation mIoU per epoch on MetalDAM (All‐three SSL vs.\ ImageNet).}
  \label{fig:uhcs-aachen-metaldam-all}
\end{figure}

\paragraph{MetalDAM.}  
The ImageNet‐pretrained encoder yields 66.73\% mIoU on MetalDAM.  MatSSL pretrained on Aachen+UHCS boosts this to 69.95\% (+3.22\%) (Table~\ref{tab:main-results}).  Figure~\ref{fig:uhcs-aachen-metaldam} shows that MatSSL converges faster and to a higher plateau than the ImageNet baseline, whereas Figure~\ref{fig:uhcs-aachen-metaldam-all} confirms similar gains when all three SSL sets are used. Compare with the state-of-the-art SSL method with the same training configuration, MatSSL surpasses DenseCL and MocoV2 by up to 1.19\% and 2.77\% mIoU, respectively.

\begin{figure}[h]
  \centering
  \includegraphics[width=1.0\linewidth]{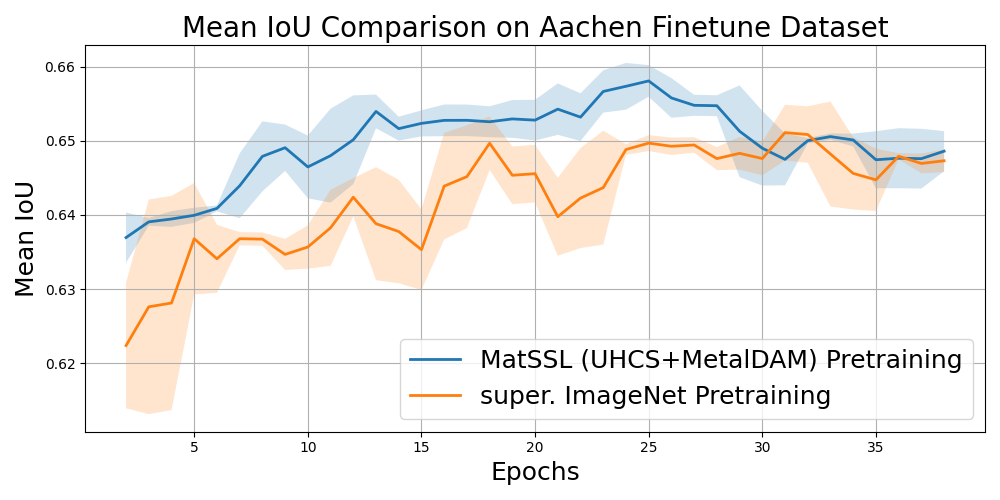}
  \caption{Validation mIoU per epoch on Aachen–Heerlen (UHCS+MetalDAM SSL vs.\ ImageNet).}
  \label{fig:uhcs-metaldam-aachen}
\end{figure}

\begin{figure}[h]
  \centering
  \includegraphics[width=1.0\linewidth]{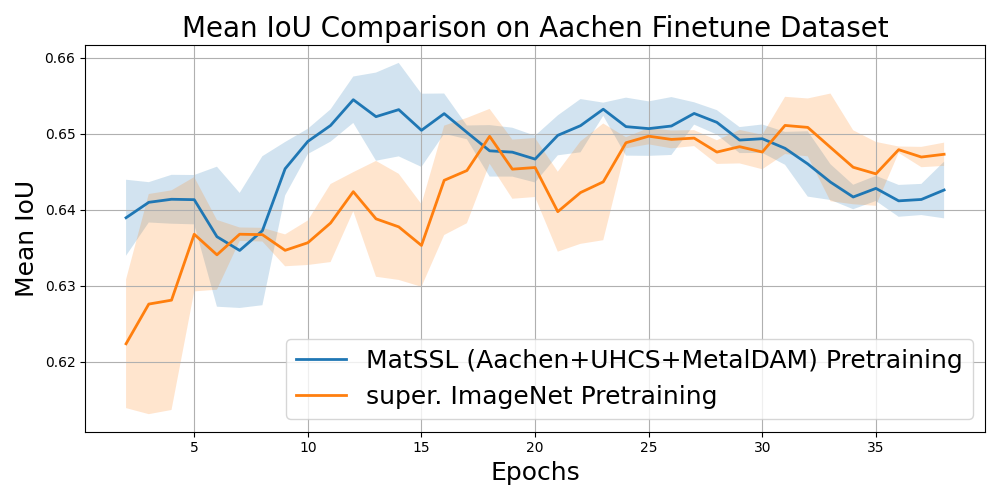}
  \caption{Validation mIoU per epoch on Aachen–Heerlen (All‐three SSL vs.\ ImageNet).}
  \label{fig:uhcs-metaldam-aachen-all}
\end{figure}

\begin{figure*}[t]
    \centering
    \includegraphics[width=1\linewidth]{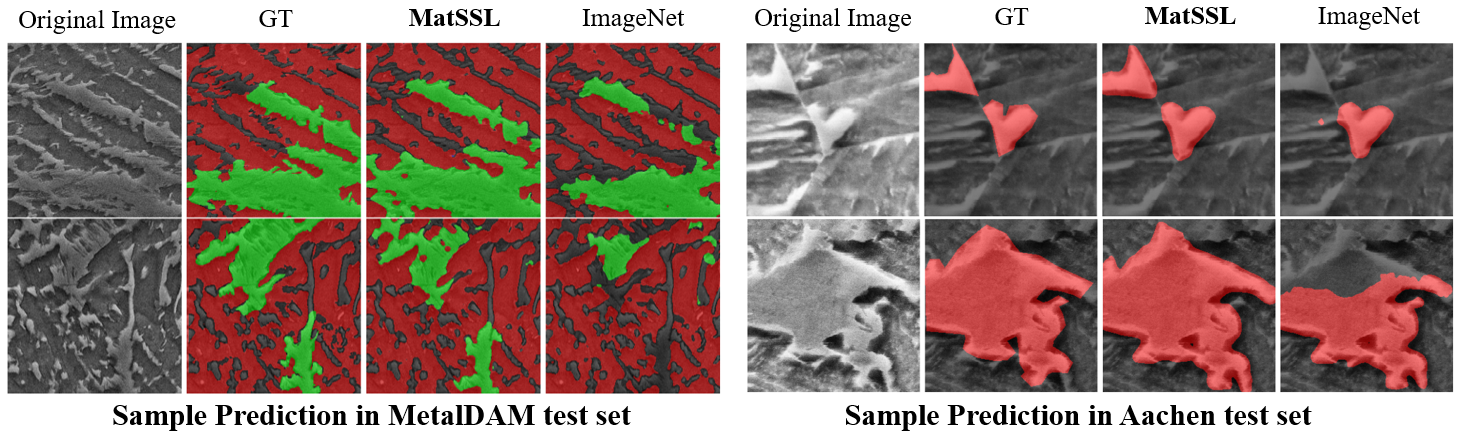}
    \caption{Sample prediction of using MatSSL as pretraining of Unet++ compared to ImageNet pretraining on MetalDAM and Aachen test set, along with Ground Truth (GT) images}
    \label{fig:sample_prediction}
\end{figure*}

\begin{table*}[!t]
    \centering
    \caption{Comparison of segmentation performance on NASA EBC benchmark test sets. MatSSL (trained on Aachen + UHCS + MetalDAM) achieves the best mIoU across all sets, with notable improvement on EBC-3.}
    \label{tab:ebc-results}
    \begin{tabular}{l l l c}
    \hline
    Finetune Dataset & Finetuning Method & Pretraining Method & Average mIoU (\%) \\
    \hline
    EBC-1 & Unet++  & MicroNet  & 95.17 \\
          & Transformer & MicroLite & 93.01 \\
          & CS-UNet& MicroNet and MicroLite & 95.98 \\
          & \textbf{Unet++} & \textbf{MatSSL (Aachen + UHCS + MetalDAM)} & \textbf{96.79} \\
    \hline
    EBC-2 & Unet++  & MicroNet  & 84.6 \\
          & Transformer & MicroLite & 84.3 \\
          & CS-UNet& MicroNet and MicroLite & 86.73 \\
          & \textbf{Unet++} & \textbf{MatSSL (Aachen + UHCS + MetalDAM)} & \textbf{94.70} \\
    \hline
    EBC-3 & Unet++  & MicroNet  & 42.58 \\
          & Transformer & MicroLite &  56.72 \\
          & CS-UNet& MicroNet and MicroLite & 45.69 \\
          & \textbf{Unet++} & \textbf{MatSSL (Aachen + UHCS + MetalDAM)} & \textbf{84.53} \\
    \hline
    
    \end{tabular}
\end{table*}

\paragraph{Aachen-Heerlen.}  
Aachen is a large annotated dataset with thousands of labeled images. The ImageNet baseline achieves 65.59\% mIoU.  MatSSL pretrained on UHCS+MetalDAM improves this to 65.98\% (+0.39\%), and adding Aachen patches to pretraining yields 65.86\%.  Both DenseCL and MoCoV2 hover around 65.8–65.9\%, showing that MatSSL’s gated fusion yields small but consistent gains (Table~\ref{tab:main-results}).  Training curves in Figure~\ref{fig:uhcs-metaldam-aachen} and Figure~\ref{fig:uhcs-metaldam-aachen-all} illustrate MatSSL’s steady advantage. The results show that MatSSL is not only effective for small or few-shot datasets but also makes improvements on a large-scale fine-tuning dataset.

Figure~\ref{fig:sample_prediction} presents a sample prediction from U-Net++ using MatSSL pretraining. This approach effectively mitigates overprediction and reduces omissions across both the Aachen and MetalDAM datasets, in contrast to the use of ImageNet pretraining.

\begin{figure*}[t]
    \centering
    \includegraphics[width=0.7\linewidth]{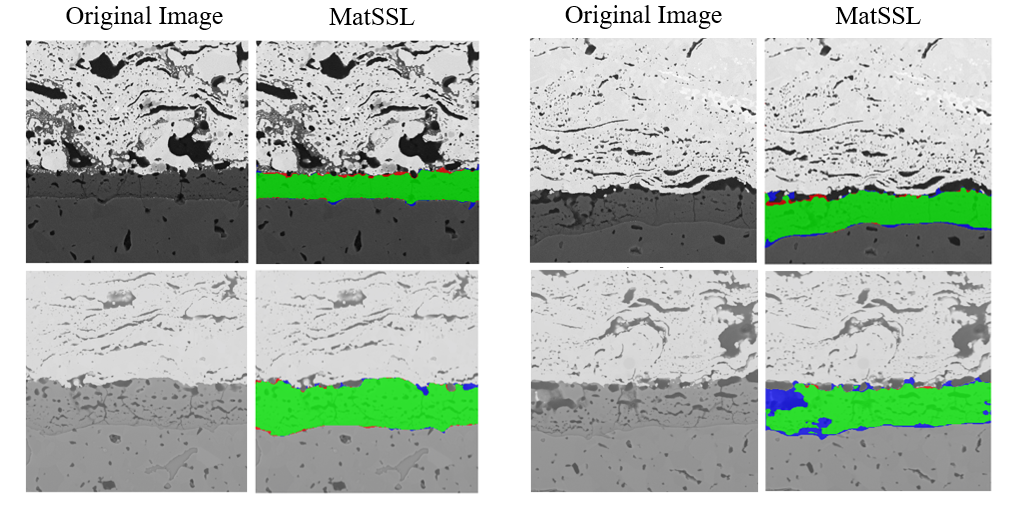}
    \caption{Visualization of sample predictions on the EBC dataset (EBC-1, EBC-2, EBC-3) using U-Net++ pretrained with MatSSL. True Positives are shown in green, False Positives in red, and False Negatives in blue.}
    \label{fig:sample_prediction_ebc}
\end{figure*}

\paragraph{Few‐Shot MicroNet (EBC).}  
Table~\ref{tab:ebc-results} summarizes the segmentation performance on the EBC benchmark across three different test splits. We compare Unet++ using MatSSL pretraining with Transformer \cite{transformer}, CS-UNet \cite{csunet}, and Unet++ using other pretraining. The MatSSL pretraining model consistently outperforms all other methods on every EBC split.

Specifically, on EBC-1, MatSSL reaches the highest average mIoU at 96.79\%, slightly surpassing the strong baseline of CS-UNet (95.98\%) and outperforming MicroNet or MicroLite pretrained baselines. For EBC-2, which features more complex damage patterns, MatSSL continues to dominate with 94.70\%, offering a significant +7.97\% gain over CS-UNet. Most notably, on the challenging EBC-3 split with only 1 training image, where other models fluctuate dramatically to average mIoUs below 57\%, MatSSL attains a remarkable 84.53\%, exceeding the next best by nearly 40 percentage points. Figure~\ref{fig:sample_prediction_ebc} illustrates a representative prediction from U-Net++ with MatSSL pretraining on the EBC dataset.

These results not only reflect the robust generalization of MatSSL across domain shifts but also highlight its stability in performance across training runs, especially in low-data and high-variance settings like EBC-3. The consistent gains up to +38.84\% mIoU on EBC-3 emphasize MatSSL’s capacity for learning highly transferable and damage-sensitive features, offering a compelling advantage for real-world structural segmentation tasks with limited supervision.

\section{Conclusion}

In this work, we propose MatSSL, a self-supervised learning (SSL) framework tailored for metallographic image segmentation in low-data regimes. By introducing a Gated Feature Fusion module into a lightweight contrastive learning pipeline, MatSSL effectively adapts a ResNet-50 encoder to the domain of metallographic micrographs without relying on large unlabeled datasets.

Extensive evaluation on real-world datasets (MetalDAM, Aachen–Heerlen) and few-shot EBC segmentation benchmarks, demonstrates the superiority of MatSSL over ImageNet-pretrained and other SSL baselines. The method achieves notable gains in segmentation accuracy, particularly in few-shot and low-annotation scenarios, affirming its robustness and generalizability.

For future work, integrating generative approaches to synthesize additional metallographic samples could further improve pretraining diversity and performance, especially in domains where acquiring annotated or even unlabeled data remains a fundamental constraint.

{
    \small
    \bibliographystyle{ieeetr}
    \bibliography{main}
}

\end{document}